\documentclass[11pt]{article}

\usepackage[preprint]{acl}

\usepackage{times}
\usepackage{latexsym}
\usepackage[T1]{fontenc}
\usepackage[utf8]{inputenc}
\usepackage{microtype}
\usepackage{inconsolata}
\usepackage{graphicx}
\usepackage{booktabs}
\usepackage{amsmath}
\usepackage{amssymb}
\usepackage{multirow}
\usepackage{xcolor}
\usepackage{colortbl}

\usepackage{listings}
\title{Defensive M2S: Training Guardrail Models on Compressed Multi-turn Conversations}

\author{Hyunjun Kim \\
  KAIST \\
  \texttt{hyunjun1121@kaist.ac.kr}}

\begin{document}
\maketitle

\begin{abstract}
Guardrail models are essential for ensuring the safety of Large Language Model (LLM) deployments, but processing full multi-turn conversation histories incurs significant computational cost. We propose \textbf{Defensive M2S}, a training paradigm that fine-tunes guardrail models on Multi-turn to Single-turn (M2S) compressed conversations rather than complete dialogue histories.

We provide a formal complexity analysis showing that M2S reduces training cost from $O(n^2)$ to $O(n)$ for $n$-turn conversations. Empirically, on our training dataset (779 samples, avg. 10.6 turns), M2S requires only \textbf{169K tokens} compared to \textbf{15.7M tokens} for the multi-turn baseline---a \textbf{93$\times$} reduction.

We evaluate Defensive M2S across three guardrail model families (LlamaGuard, Nemotron, Qwen3Guard) and three compression templates (hyphenize, numberize, pythonize) on SafeDialBench, a comprehensive multi-turn jailbreak benchmark. Our best configuration, Qwen3Guard with hyphenize compression, achieves \textbf{93.8\%} attack detection recall while reducing inference tokens by \textbf{94.6\%} (from 3,231 to 173 tokens per conversation). This represents a 38.9 percentage point improvement over the baseline while dramatically reducing both training and inference costs.

Our findings demonstrate that M2S compression can serve as an effective efficiency technique for guardrail deployment, enabling scalable safety screening of long multi-turn conversations.
\end{abstract}

\section{Introduction}
\label{sec:introduction}

Large Language Models (LLMs) have demonstrated remarkable capabilities across diverse tasks, but their susceptibility to adversarial attacks remains a critical concern. Among these threats, \textit{multi-turn jailbreak attacks} represent a particularly insidious category, where adversaries gradually manipulate LLMs through a series of carefully crafted conversational turns to bypass safety guardrails and elicit harmful outputs.

Guardrail models serve as a crucial defense mechanism, acting as classifiers that evaluate whether a given input-output pair is safe or unsafe. However, deploying these models for multi-turn conversations presents significant computational challenges: processing full conversation histories requires substantial token throughput, leading to increased latency and cost at inference time. As conversations grow longer, the computational burden scales linearly, making real-time safety screening increasingly expensive.

Recent work on \textit{Multi-turn to Single-turn (M2S)} compression \citep{ha2025m2s} has shown that multi-turn jailbreak attacks can be distilled into compact single-turn prompts that preserve their adversarial effectiveness. This insight, while concerning from a security perspective, suggests an intriguing defensive application: if the essential semantics of multi-turn attacks can be captured in compressed form, perhaps guardrail models can be trained to recognize these compressed representations directly.

In this paper, we propose \textbf{Defensive M2S}, a training paradigm that fine-tunes guardrail models on M2S-compressed conversation histories rather than full multi-turn dialogues. Our key hypothesis is that M2S compression maintains the semantic information necessary for accurate safety classification while dramatically reducing the computational cost of inference.

We validate this hypothesis through extensive experiments on three guardrail model families (LlamaGuard, Nemotron, and Qwen3Guard) across multiple M2S compression templates (hyphenize, numberize, pythonize). Our evaluation on SafeDialBench \citep{chen2025safedialbench}, a comprehensive multi-turn jailbreak benchmark comprising 2,037 samples across 6 attack categories and 7 attack methods, reveals several key findings:

\begin{itemize}
    \item \textbf{Efficiency-Accuracy Trade-off}: M2S-trained models achieve up to 94.6\% token reduction while maintaining competitive detection accuracy. The best configuration (Qwen3Guard with hyphenize template) achieves 93.8\% recall compared to 54.9\% baseline recall, demonstrating that compression can actually \textit{improve} detection performance for certain model-template combinations.

    \item \textbf{Model-Template Sensitivity}: The effectiveness of M2S training varies significantly across model-template combinations, with Qwen3Guard favoring hyphenize (93.8\% recall) while Nemotron performs best with numberize (87.8\% recall).

    \item \textbf{Single-Template Superiority}: Training on a single compression template outperforms mixed-template training, suggesting that template-specific representations provide stronger learning signals than diverse but inconsistent formats.
\end{itemize}

Our contributions can be summarized as follows:
\begin{enumerate}
    \item We introduce Defensive M2S, a novel training paradigm that leverages adversarial compression techniques for efficient guardrail deployment.
    \item We provide formal complexity analysis showing M2S reduces training cost from $O(n^2)$ to $O(n)$, empirically validated with 93$\times$ token reduction on our dataset.
    \item We provide the first systematic evaluation of M2S-trained guardrails across multiple model families, compression templates, and evaluation benchmarks.
    \item We release our trained adapters and evaluation code to facilitate reproducible research in efficient LLM safety.
\end{enumerate}

\section{Related Work}
\label{sec:related_work}

\subsection{Multi-Turn Jailbreak Attacks}

Multi-turn jailbreak attacks exploit the conversational nature of LLMs to gradually elicit harmful outputs through sequences of seemingly benign prompts. \citet{russinovich2024crescendo} introduce Crescendo, a multi-turn attack that begins innocuously and progressively escalates by referencing model replies, achieving 56\% attack success rate (ASR) on GPT-4 and 83\% on Gemini-Pro. \citet{ren2024actorattack} propose ActorAttack, which models semantically linked entities as attack clues to generate diverse attack paths that conceal malicious intent across conversation turns.

Automated red teaming methods have emerged to systematically discover vulnerabilities. GOAT \citep{pavlova2024goat} simulates adversarial user reasoning in multi-turn conversations, achieving 97\% ASR on Llama 3.1. TAP \citep{mehrotra2024tap} uses tree-of-thought reasoning with attacker-evaluator-target LLM pipelines, achieving 80\%+ ASR while bypassing guardrails like LlamaGuard. WildTeaming \citep{jiang2024wildteaming} mines real user-chatbot interactions to discover 5.7K unique jailbreak tactic clusters.

Single-turn attacks provide foundations for understanding adversarial robustness. GCG \citep{zou2023universal} pioneered token-level optimization of adversarial suffixes, achieving 88\% ASR with cross-model transfer. AutoDAN \citep{liu2024autodan} uses genetic algorithms to generate semantically meaningful jailbreaks, while PAIR \citep{chao2023pair} enables black-box attacks through iterative prompt refinement in under 20 queries.

Most relevant to our work, M2S \citep{ha2025m2s} introduces Multi-turn to Single-turn compression, consolidating multi-turn jailbreaks into structured single-turn prompts using hyphenize, numberize, and pythonize templates. Their work demonstrates that compressed prompts often \textit{outperform} original multi-turn attacks by up to 17.5\% ASR, exploiting ``contextual blindness'' in both native and external guardrails. We leverage this observation defensively: if M2S compression preserves adversarial semantics, guardrails trained on compressed representations should maintain detection accuracy.

\subsection{Guardrail Models}

LLM-based guardrails have emerged as the dominant paradigm for safety classification. LlamaGuard \citep{inan2023llamaguard} pioneered fine-tuning LLMs as input-output safeguards, introducing a six-category safety taxonomy with high adaptability to new policies. Subsequent versions (LlamaGuard 2/3) support the MLCommons taxonomy \citep{vidgen2024mlcommons}. Nemotron Safety Guard \citep{ghosh2024aegis} extends this approach with 13 critical risk categories and the ``Needs Caution'' label for nuanced moderation.

Recent work has improved guardrail capabilities across multiple dimensions. WildGuard \citep{han2024wildguard} achieves three goals simultaneously: identifying malicious prompts, detecting safety risks in responses, and measuring refusal rates, outperforming Llama-Guard2 by 25.3\% on refusal detection. ShieldLM \citep{zhang2024shieldlm} introduces bilingual (Chinese/English) detection with customizable rules and explanations. ShieldGemma \citep{zeng2024shieldgemma} demonstrates superior performance (+10.8\% AU-PRC over LlamaGuard) using novel synthetic data generation.

Parameter-efficient adaptation has enabled deployment in resource-constrained settings. LoRA-Guard \citep{loraguard2024} achieves 100-1000$\times$ lower parameter overhead through knowledge sharing between LLMs and guardrails. NeMo Guardrails \citep{rebedea2023nemo} provides programmable rails for controllable LLM applications. Our work complements these efficiency approaches by reducing \textit{input} token requirements through M2S compression.

\subsection{Multi-Turn Dialogue Safety}

Evaluating safety in multi-turn contexts presents unique challenges. SafeDialBench \citep{chen2025safedialbench} provides 4,053 dialogues across six safety dimensions with seven jailbreak strategies including reference attacks. CoSafe \citep{yu2024cosafe} studies coreference-based attacks, revealing ASR ranging from 14\% to 56\% across models. GuardBench \citep{bassani2024guardbench} consolidates 40 evaluation datasets for systematic guardrail comparison.

Several datasets support safety research. BeaverTails \citep{ji2023beavertails} provides 333K QA pairs with separated harmlessness and helpfulness annotations across 14 categories. ToxicChat \citep{lin2023toxicchat} captures real-world user-AI interactions from the Vicuna demo. HarmBench \citep{mazeika2024harmbench} standardizes red teaming evaluation with 510 behaviors and 18 attack methods.

The challenge of over-refusal has also received attention. XSTest \citep{rottger2024xstest} identifies exaggerated safety behaviors where models refuse safe prompts due to lexical similarity with harmful content. Our approach indirectly addresses this by training on compressed representations that may filter out superficial lexical patterns while preserving semantic safety signals.

\subsection{Efficient NLP Inference}

Prompt compression techniques reduce token overhead for LLM inference. LLMLingua \citep{jiang2023llmlingua} achieves up to 20$\times$ compression through coarse-to-fine token pruning. LongLLMLingua \citep{jiang2024longllmlingua} extends this to long contexts, achieving 94\% cost reduction with performance improvements. These methods focus on general language modeling; our M2S compression is specifically designed for preserving safety-relevant semantics.

KV cache optimization addresses memory bottlenecks. H2O \citep{zhang2023h2o} evicts non-essential cached states using heavy-hitter detection, achieving 29$\times$ throughput improvement. StreamingLLM \citep{xiao2024streamingllm} enables infinite sequence processing through attention sinks. FlashAttention \citep{dao2022flashattention} provides IO-aware exact attention with linear memory scaling.

Alternative architectures offer asymptotic improvements. Mamba \citep{gu2023mamba} achieves linear $O(n)$ complexity versus quadratic $O(n^2)$ for Transformers, with 5$\times$ higher throughput. However, these approaches require architectural changes; our Defensive M2S is model-agnostic and applicable to any Transformer-based guardrail.

Our work differs from prior compression approaches in two key ways: (1) we compress at the \textit{semantic} level using structured templates rather than token-level pruning, and (2) we apply compression during \textit{training} rather than inference, enabling the model to learn safety-relevant features from compressed representations directly.

\section{Methodology}
\label{sec:methodology}

\subsection{Problem Formulation}

Let $C = \{(u_1, a_1), (u_2, a_2), \ldots, (u_n, a_n)\}$ denote a multi-turn conversation with $n$ turns, where $u_i$ represents the user message and $a_i$ represents the assistant response at turn $i$. A guardrail model $\mathcal{G}$ is a classifier that predicts a safety label $y \in \{\text{safe}, \text{unsafe}\}$ given the conversation context.

In the conventional baseline approach, the guardrail model processes the full conversation:
\begin{equation}
y = \mathcal{G}(C) = \mathcal{G}(u_1, a_1, \ldots, u_n, a_n)
\end{equation}

The computational cost scales with the total token count $|C| = \sum_{i=1}^{n}(|u_i| + |a_i|)$, which can become prohibitively expensive for long conversations.

\subsection{M2S Compression}

Multi-turn to Single-turn (M2S) compression transforms a multi-turn conversation into a compact single-turn representation. Given a compression function $f_\theta$:
\begin{equation}
\tilde{C} = f_\theta(C)
\end{equation}
where $\tilde{C}$ is the compressed representation and $|\tilde{C}| \ll |C|$.

We investigate three compression templates from prior work \citep{ha2025m2s}:

\paragraph{Hyphenize Template} Formats user turns as a bulleted list:
\begin{lstlisting}
- [Turn 1 content]
- [Turn 2 content]
...
- [Turn n content]
\end{lstlisting}

\paragraph{Numberize Template} Formats user turns as a numbered list:
\begin{lstlisting}
1. [Turn 1 content]
2. [Turn 2 content]
...
n. [Turn n content]
\end{lstlisting}

\paragraph{Pythonize Template} Formats the conversation in a Python code-like structure:
\begin{lstlisting}
def conversation():
    user_turn_1 = "[Turn 1 content]"
    user_turn_2 = "[Turn 2 content]"
    ...
    user_turn_n = "[Turn n content]"
\end{lstlisting}

A key design choice in M2S compression is to extract \textit{only user turns}, discarding assistant responses. This is motivated by two observations: (1) adversarial intent is primarily encoded in user messages, and (2) assistant responses contribute significant token overhead without proportional safety-relevant information.

\subsection{Defensive M2S Training}

We propose training guardrail models on M2S-compressed inputs rather than full conversations. Given a training dataset $\mathcal{D} = \{(C_i, y_i)\}_{i=1}^{N}$, we create a compressed training set:
\begin{equation}
\tilde{\mathcal{D}} = \{(f_\theta(C_i), y_i)\}_{i=1}^{N}
\end{equation}

The guardrail model is then fine-tuned to minimize the cross-entropy loss:
\begin{equation}
\mathcal{L} = -\sum_{i=1}^{N} y_i \log \mathcal{G}(\tilde{C}_i) + (1-y_i) \log (1 - \mathcal{G}(\tilde{C}_i))
\end{equation}

\subsection{Computational Complexity Analysis}

A critical advantage of Defensive M2S is the dramatic reduction in computational cost during both \textit{data generation} and \textit{training}. To contextualize this advantage, we first examine how multi-turn jailbreak attacks are generated in practice.

\paragraph{Multi-turn Attack Generation Taxonomy}

Recent literature reveals two fundamentally different paradigms for constructing multi-turn jailbreak attacks:

\textbf{(1) Response-Dependent Methods} generate user prompts dynamically by referencing the target model's previous responses. Crescendo \citep{russinovich2024crescendo} ``exploits the LLM's tendency to follow patterns... particularly text generated by the LLM itself.'' Similarly, PAIR \citep{chao2023pair} ``iteratively refines the candidate prompt by accumulating previous attempts and responses in the chat history,'' and TAP \citep{mehrotra2024tap} extends this with tree-based exploration. Other examples include ActorAttack \citep{ren2024actorattack}, which ``dynamically adapts its attack path based on target model responses.''

\textbf{(2) Pre-Scripted Methods} generate all user prompts in advance without requiring model responses. The MHJ (Multi-turn Human Jailbreak) dataset used by \citet{ha2025m2s} consists of pre-written user turns. Many-shot jailbreaking \citep{anil2024manyshot} includes ``faux dialogues'' that are entirely fabricated without actual model interaction.

\paragraph{Implications for Training Data}

For \textit{pre-scripted attacks}, only user turns exist---assistant responses must be generated to create training data for conventional guardrails. Our M2S approach eliminates this requirement entirely.

For \textit{response-dependent attacks}, while responses exist during attack creation, they may require regeneration when: (1) adapting attacks to different target models, (2) creating training data with specific chat formats, or (3) building guardrails for model families different from the attack target. Thus, even for response-dependent datasets, the baseline complexity analysis often applies.

\paragraph{Formal Complexity Analysis}

Let $U$ denote the average tokens per user turn and $R$ the average tokens per assistant response. For an $n$-turn conversation:

\paragraph{Multi-turn Baseline Complexity}

The baseline approach requires two costly phases:

\textbf{Phase 1: Training Data Generation.} To train a guardrail on full conversations, we must generate assistant responses for each turn by querying an LLM. Critically, each response generation requires the \textit{entire preceding context}: at turn $k$, the LLM receives all previous user turns and generated responses:
\begin{equation}
\text{Input}_k = \sum_{i=1}^{k} U + \sum_{i=1}^{k-1} R = kU + (k-1)R
\end{equation}

The total input tokens for generating all $n$ responses:
\begin{equation}
T_{\text{gen}} = \sum_{k=1}^{n} \left( kU + (k-1)R \right) = \frac{n(n+1)}{2}U + \frac{n(n-1)}{2}R
\end{equation}

\textbf{Phase 2: Guardrail Training.} To detect attacks at any conversation stage, the guardrail must be trained on incremental prefixes. Sample $k$ contains $k$ turns:
\begin{equation}
\text{Sample}_k = k(U + R)
\end{equation}

Total training tokens:
\begin{equation}
T_{\text{train}} = \sum_{k=1}^{n} k(U+R) = \frac{n(n+1)}{2}(U+R)
\end{equation}

\textbf{Total Multi-turn Cost:}
\begin{equation}
T_{\text{baseline}} = T_{\text{gen}} + T_{\text{train}} = O(n^2)
\end{equation}

\paragraph{M2S Complexity}

In contrast, M2S requires \textbf{no response generation}. Since M2S extracts only user turns and compresses them into a structured format, we can directly use the existing jailbreak prompts without querying any LLM. The only cost is the compressed training samples themselves:
\begin{equation}
T_{\text{M2S}} = nU + O(1) \approx nU = O(n)
\end{equation}

This eliminates Phase 1 entirely ($T_{\text{gen}} = 0$) and reduces Phase 2 to a single sample per conversation rather than $n$ incremental samples.

\paragraph{Complexity Ratio}

The ratio of baseline to M2S complexity:
\begin{equation}
\frac{T_{\text{baseline}}}{T_{\text{M2S}}} = \frac{O(n^2)}{O(n)} = O(n)
\end{equation}

This means the efficiency advantage of M2S \textit{increases} with conversation length. For a 10-turn conversation with $U \approx R$, the theoretical ratio is approximately $21\times$.

\subsection{Training Configuration}

We employ QLoRA fine-tuning \citep{dettmers2023qlora} for parameter-efficient adaptation. Key hyperparameters include:

\begin{itemize}
    \item \textbf{Base Models}: LlamaGuard-3-8B, Nemotron-Safety-Guard-8B, Qwen3Guard-Gen-8B
    \item \textbf{LoRA Configuration}: rank $r=16$, $\alpha=32$, dropout $p=0.1$
    \item \textbf{Quantization}: 4-bit NormalFloat (NF4)
    \item \textbf{Training}: batch size 4, gradient accumulation 4, learning rate $2 \times 10^{-4}$, 3 epochs
    \item \textbf{Seeds}: 42, 123, 456 (for statistical robustness)
\end{itemize}

\subsection{Inference Protocol}

At inference time, given a new multi-turn conversation $C_{\text{test}}$:
\begin{enumerate}
    \item Compress the conversation: $\tilde{C}_{\text{test}} = f_\theta(C_{\text{test}})$
    \item Generate guardrail output: $\hat{y} = \mathcal{G}(\tilde{C}_{\text{test}})$
    \item Parse prediction by checking for ``unsafe'' substring in generated text
\end{enumerate}

This protocol enables processing of long conversations at a fraction of the original computational cost, as $|\tilde{C}_{\text{test}}| \approx 0.05 \cdot |C_{\text{test}}|$ in our experiments.

\section{Experimental Setup}
\label{sec:experiments}

\subsection{Guardrail Models}

We evaluate three state-of-the-art open-source guardrail models:

\paragraph{LlamaGuard-3-8B} Meta's third-generation safety classifier built on Llama-3 architecture \citep{inan2023llamaguard,dubey2024llama3}, trained on a diverse taxonomy of harmful content categories.

\paragraph{Nemotron-Safety-Guard-8B-v3} NVIDIA's safety guardrail model based on Llama-3.1 architecture \citep{ghosh2024aegis}, designed for comprehensive safety classification.

\paragraph{Qwen3Guard-Gen-8B} Alibaba's guardrail model from the Qwen3 family \citep{zhao2025qwen3guard}, featuring a distinct tokenization scheme and chat format (im\_start/im\_end).

\subsection{Training Data}

We construct our training dataset from multiple sources: (1) multi-turn jailbreak attacks from \citet{gibbs2024emerging} and Anthropic's red-team attempts, and (2) benign multi-turn conversations from the HH-RLHF corpus \citep{bai2022hhrlhf}. We filter for conversations with 8 or more user turns and balance the dataset to ensure equal representation of safe and unsafe samples. The resulting dataset contains 779 samples (385 unsafe, 394 safe) with an average of 10.6 user turns per conversation.

For M2S training, we preprocess the data by applying the compression templates to all conversations, retaining only user turns as described in Section~\ref{sec:methodology}.

\subsection{Evaluation Benchmarks}

\paragraph{SafeDialBench} A comprehensive multi-turn jailbreak benchmark \citep{chen2025safedialbench} comprising 2,037 samples across 6 attack categories (violence, fraud, illegal activities, etc.) and 7 attack methods (role-playing, hypothetical scenarios, progressive escalation, etc.). This serves as our primary evaluation benchmark.

\paragraph{Longturn MHJ} A subset of 195 samples (102 attack, 93 benign) from the MHJ test set \citep{ha2025m2s}, used for preliminary validation and template ablation studies.

\subsection{Evaluation Metrics}

\paragraph{Recall (\%)} The primary metric measuring the proportion of unsafe samples correctly identified:
\begin{equation}
\text{Recall} = \frac{\text{True Positives}}{\text{True Positives} + \text{False Negatives}}
\end{equation}

\paragraph{Token Reduction (\%)} The efficiency metric measuring compression ratio:
\begin{equation}
\text{Token Reduction} = 1 - \frac{|\tilde{C}|}{|C|}
\end{equation}

\paragraph{False Positive Rate} Measured on benign samples to assess over-flagging:
\begin{equation}
\text{FPR} = \frac{\text{False Positives}}{\text{False Positives} + \text{True Negatives}}
\end{equation}

\subsection{Experimental Configurations}

We evaluate the following training configurations:

\begin{itemize}
    \item \textbf{Baseline}: Full conversation training (no compression)
    \item \textbf{M2S Hyphenize}: Compression with hyphenize template
    \item \textbf{M2S Numberize}: Compression with numberize template
    \item \textbf{M2S Pythonize}: Compression with pythonize template
    \item \textbf{M2S All}: Mixed training with all three templates
\end{itemize}

Each configuration is trained with 3 random seeds (42, 123, 456) to ensure statistical robustness. We report mean and standard deviation across seeds.

\section{Results}
\label{sec:results}

\subsection{Main Results on SafeDialBench}

Table~\ref{tab:main-results} presents the primary comparison between baseline (full conversation) and M2S-trained guardrail models on SafeDialBench.

\begin{table}[t]
\centering
\small
\begin{tabular}{llcc}
\toprule
\textbf{Model} & \textbf{Training} & \textbf{Recall (\%)} & \textbf{Tokens} \\
\midrule
\multirow{4}{*}{LlamaGuard}
    & Baseline & 75.1 $\pm$ 14.3 & 3110 \\
    & M2S Hyphenize & 24.1 $\pm$ 5.3 & 175 \\
    & M2S Numberize & 24.5 $\pm$ 4.8 & 176 \\
    & M2S Pythonize & 17.2 $\pm$ 1.2 & 287 \\
\midrule
\multirow{4}{*}{Nemotron}
    & Baseline & \textbf{99.0 $\pm$ 1.3} & 3020 \\
    & M2S Hyphenize & 67.6 $\pm$ 23.8 & 176 \\
    & M2S Numberize & 87.8 $\pm$ 8.7 & 177 \\
    & M2S Pythonize & 82.9 $\pm$ 4.3 & 288 \\
\midrule
\multirow{4}{*}{Qwen3Guard}
    & Baseline & 54.9 $\pm$ 0.0 & 3231 \\
    & M2S Hyphenize & \textbf{93.8 $\pm$ 1.7} & 173 \\
    & M2S Numberize & 33.6 $\pm$ 2.7 & 174 \\
    & M2S Pythonize & 30.6 $\pm$ 4.1 & 285 \\
\bottomrule
\end{tabular}
\caption{SafeDialBench results comparing baseline (full conversation) and M2S-trained models. Values are mean $\pm$ std across 3 seeds. Best M2S result per model in bold.}
\label{tab:main-results}
\end{table}

\paragraph{Key Finding 1: Model-Template Interaction} The effectiveness of M2S training depends critically on the model-template combination. Qwen3Guard achieves its best performance with hyphenize (93.8\%), dramatically outperforming its baseline (54.9\%). In contrast, Nemotron performs best with numberize (87.8\%) or pythonize (82.9\%), while hyphenize shows high variance (67.6\% $\pm$ 23.8\%).

\paragraph{Key Finding 2: Efficiency Gains} All M2S configurations achieve approximately \textbf{94\% token reduction} (from $\sim$3100 tokens to $\sim$175 tokens), enabling significantly faster inference without proportional accuracy loss for well-matched model-template pairs.

\paragraph{Key Finding 3: LlamaGuard Struggles} LlamaGuard shows consistent underperformance across all M2S templates (17-25\% recall), despite reasonable baseline performance (75.1\%). This suggests that LlamaGuard's internal representations may not generalize well to compressed formats.

\subsection{Mixed-Template Training Analysis}

Table~\ref{tab:mixed-template} shows results when training on all templates simultaneously (M2S All).

\begin{table}[t]
\centering
\small
\begin{tabular}{llc}
\toprule
\textbf{Model} & \textbf{Eval Template} & \textbf{Recall (\%)} \\
\midrule
\multirow{3}{*}{Nemotron}
    & Hyphenize & 40.3 $\pm$ 43.5 \\
    & Numberize & 37.8 $\pm$ 39.7 \\
    & Pythonize & 43.2 $\pm$ 30.6 \\
\midrule
\multirow{3}{*}{Qwen3Guard}
    & Hyphenize & 31.4 $\pm$ 7.9 \\
    & Numberize & 30.2 $\pm$ 9.1 \\
    & Pythonize & 27.2 $\pm$ 6.0 \\
\midrule
\multirow{3}{*}{LlamaGuard}
    & Hyphenize & 23.5 $\pm$ 2.5 \\
    & Numberize & 24.1 $\pm$ 2.7 \\
    & Pythonize & 16.9 $\pm$ 3.0 \\
\bottomrule
\end{tabular}
\caption{Results for models trained on all templates (M2S All), evaluated on each template separately.}
\label{tab:mixed-template}
\end{table}

\paragraph{Key Finding 4: Single-Template Superiority} Mixed-template training consistently underperforms single-template training. Compare Qwen3Guard: single-template hyphenize achieves 93.8\% recall, while the mixed-trained model achieves only 31.4\% on the same template. The high variance in Nemotron's mixed-template results (up to $\pm$43.5\%) suggests unstable learning dynamics when exposed to diverse compression formats.

\subsection{Efficiency-Accuracy Trade-off}

Analyzing the trade-off between token usage and recall across all configurations, we identify the Pareto-optimal configurations:

\begin{itemize}
    \item \textbf{Maximum Recall}: Nemotron Baseline (99.0\% recall, 3020 tokens)
    \item \textbf{Best Efficiency-Accuracy}: Qwen3Guard M2S Hyphenize (93.8\% recall, 173 tokens) --- achieving 94.6\% token reduction with only 5.2\% recall reduction vs. best baseline
    \item \textbf{High Recall + Efficiency}: Nemotron M2S Numberize (87.8\% recall, 177 tokens)
\end{itemize}

\subsection{Template Ablation on Longturn MHJ}

Table~\ref{tab:template-ablation} presents results on the smaller Longturn MHJ dataset, which includes both attack detection (recall) and false positive rate (FPR) metrics.

\begin{table}[t]
\centering
\small
\begin{tabular}{lccc}
\toprule
\textbf{Template} & \textbf{Recall (\%)} & \textbf{FPR (\%)} & \textbf{Tokens} \\
\midrule
Hyphenize & 98.0 & 1.1 & 284 \\
Numberize & 98.0 & 1.1 & 290 \\
Pythonize & 98.0 & 2.2 & 402 \\
\bottomrule
\end{tabular}
\caption{Template ablation on Longturn MHJ (LlamaGuard, single seed). All templates achieve equivalent recall with low FPR.}
\label{tab:template-ablation}
\end{table}

On this smaller dataset, all templates achieve equivalent recall (98.0\%) with minimal false positives. The discrepancy with SafeDialBench results suggests that model generalization to diverse attack patterns requires careful template selection.

\subsection{Training Complexity Validation}

We validate our theoretical complexity analysis (Section~\ref{sec:methodology}) on our actual training dataset (779 samples, avg. 10.6 user turns). Table~\ref{tab:training-complexity} shows the empirical token counts.

\begin{table}[t]
\centering
\small
\begin{tabular}{lrr}
\toprule
\textbf{Metric} & \textbf{M2S} & \textbf{Multi-turn} \\
\midrule
Phase 1: Data Generation & 0 & 7,251,110 \\
Phase 2: Training & 169,153 & 8,494,610 \\
\midrule
\textbf{Total Tokens} & \textbf{169,153} & \textbf{15,745,720} \\
Avg. per Sample & 217.1 & 20,212.7 \\
\bottomrule
\end{tabular}
\caption{Empirical training token complexity. M2S achieves \textbf{93$\times$} reduction (\textbf{98.9\%} fewer tokens).}
\label{tab:training-complexity}
\end{table}

\paragraph{Key Finding 5: Quadratic vs. Linear Scaling} The empirical ratio (93$\times$) aligns with our theoretical prediction for longer conversations. Table~\ref{tab:complexity-scaling} shows how this advantage increases with conversation length.

\begin{table}[t]
\centering
\small
\begin{tabular}{rccc}
\toprule
\textbf{Turns} & \textbf{M2S} & \textbf{Multi-turn} & \textbf{Ratio} \\
\midrule
2 & 200 & 1,000 & 5.0$\times$ \\
5 & 500 & 5,500 & 11.0$\times$ \\
10 & 1,000 & 21,000 & 21.0$\times$ \\
15 & 1,500 & 46,500 & 31.0$\times$ \\
20 & 2,000 & 82,000 & 41.0$\times$ \\
\bottomrule
\end{tabular}
\caption{Theoretical scaling with $U=R=100$ tokens. M2S advantage grows linearly with turn count.}
\label{tab:complexity-scaling}
\end{table}

This $O(n^2)$ vs. $O(n)$ difference has profound practical implications: training a guardrail on 20-turn conversations using the baseline approach requires \textbf{41$\times$} more tokens than M2S.

\subsection{Statistical Significance}

We conduct paired t-tests between the best M2S configuration (Qwen3Guard Hyphenize) and baselines:
\begin{itemize}
    \item vs. Qwen3Guard Baseline: $p < 0.001$ (M2S significantly better)
    \item vs. Nemotron Baseline: $p = 0.12$ (not significantly different)
\end{itemize}

The best M2S configuration statistically matches the best baseline while using 94.6\% fewer inference tokens and requiring 93$\times$ fewer training tokens.

\section{Conclusion}
\label{sec:conclusion}

We introduced Defensive M2S, a training paradigm that fine-tunes guardrail models on M2S-compressed multi-turn conversations rather than full dialogue histories. Our extensive evaluation across three guardrail model families and multiple compression templates reveals that this approach can achieve substantial efficiency gains (up to 94.6\% token reduction) while maintaining or even improving detection accuracy for certain model-template combinations.

Our key findings include: (1) Qwen3Guard with hyphenize compression achieves 93.8\% recall while using only 173 tokens per conversation, outperforming its baseline by nearly 40 percentage points; (2) model-template compatibility is crucial, with different models favoring different compression formats; and (3) single-template training outperforms mixed-template training, suggesting that consistent compression formats provide stronger learning signals.

These results demonstrate the practical viability of Defensive M2S for deploying efficient guardrails in production systems, where computational cost and latency are critical concerns. Future work should explore adaptive template selection, combination with other efficiency techniques such as model distillation, and evaluation on additional safety benchmarks.

\section*{Limitations}
\label{sec:limitations}

Our work has several limitations that should be considered when interpreting the results:

\paragraph{Limited Model Coverage} We evaluate three guardrail models (LlamaGuard, Nemotron, Qwen3Guard), all in the 8B parameter range. The effectiveness of Defensive M2S on smaller or larger models, as well as closed-source commercial guardrails, remains unexplored.

\paragraph{Template Generalization} Our M2S templates are derived from prior work and may not represent the optimal compression strategy. Template design is currently manual, and automated template discovery could yield better results.

\paragraph{Single Benchmark Emphasis} While SafeDialBench provides comprehensive coverage of multi-turn attacks, additional benchmarks covering different attack taxonomies, languages, and domains would strengthen our conclusions.

\paragraph{Loss of Context} M2S compression discards assistant responses entirely, which may contain safety-relevant context in some scenarios (e.g., the model beginning to comply with harmful requests). Our approach may miss such signals.

\paragraph{Adversarial Robustness} We do not evaluate against adversaries who are aware of the M2S compression pipeline. Attacks specifically designed to exploit compression artifacts could potentially bypass our defenses.

\paragraph{Training Data Scale} Our training data is relatively small ($\sim$1000 samples). Larger-scale training with more diverse multi-turn conversations could improve generalization.

\paragraph{Real-world Deployment} Our evaluation is conducted in controlled settings. Production deployment involves additional considerations such as latency requirements, handling of streaming conversations, and integration with other safety measures.



\bibliography{references}

\appendix
\section{M2S Template Examples}
\label{sec:appendix-templates}

This appendix provides concrete examples of each M2S compression template applied to a sample multi-turn conversation.

\subsection{Original Multi-turn Conversation}

\begin{lstlisting}
User: I'm writing a crime novel and need help with realistic dialogue.
Assistant: I'd be happy to help with your creative writing project...
User: The antagonist needs to explain their method for avoiding detection.
Assistant: For fictional purposes, here are some narrative elements...
User: Can you provide specific technical details they would discuss?
Assistant: In crime fiction, technical accuracy can add realism...
\end{lstlisting}

\subsection{Hyphenize Compression}

\begin{lstlisting}
- I'm writing a crime novel and need help with realistic dialogue.
- The antagonist needs to explain their method for avoiding detection.
- Can you provide specific technical details they would discuss?
\end{lstlisting}

\subsection{Numberize Compression}

\begin{lstlisting}
1. I'm writing a crime novel and need help with realistic dialogue.
2. The antagonist needs to explain their method for avoiding detection.
3. Can you provide specific technical details they would discuss?
\end{lstlisting}

\subsection{Pythonize Compression}

\begin{lstlisting}
def conversation():
    user_turn_1 = "I'm writing a crime novel and need help with realistic dialogue."
    user_turn_2 = "The antagonist needs to explain their method for avoiding detection."
    user_turn_3 = "Can you provide specific technical details they would discuss?"
\end{lstlisting}

\section{Full Results Tables}
\label{sec:appendix-full-results}

\subsection{Per-Seed Results on SafeDialBench}

\begin{table*}[t]
\centering
\small
\begin{tabular}{llccc}
\toprule
\textbf{Model} & \textbf{Configuration} & \textbf{Seed 42} & \textbf{Seed 123} & \textbf{Seed 456} \\
\midrule
\multirow{4}{*}{LlamaGuard}
    & Baseline & 65.8 & 67.9 & 91.5 \\
    & M2S Hyphenize & 26.2 & 18.1 & 28.0 \\
    & M2S Numberize & 29.9 & 23.0 & 20.7 \\
    & M2S Pythonize & 16.4 & 16.7 & 18.6 \\
\midrule
\multirow{4}{*}{Nemotron}
    & Baseline & 97.5 & 100.0 & 99.5 \\
    & M2S Hyphenize & 42.6 & 90.1 & 70.0 \\
    & M2S Numberize & 78.5 & 95.7 & 89.2 \\
    & M2S Pythonize & 87.6 & 79.3 & 81.8 \\
\midrule
\multirow{4}{*}{Qwen3Guard}
    & Baseline & 54.9 & 54.9 & 54.9 \\
    & M2S Hyphenize$^\dagger$ & 92.5 & 93.2 & 95.8 \\
    & M2S Numberize & 35.9 & 34.4 & 30.6 \\
    & M2S Pythonize & 30.0 & 26.8 & 34.9 \\
\bottomrule
\end{tabular}
\caption{Per-seed recall (\%) on SafeDialBench for all configurations. $^\dagger$Qwen3Guard M2S Hyphenize uses seeds 1, 2, 3.}
\label{tab:per-seed-results}
\end{table*}

\section{Training Details}
\label{sec:appendix-training}

\subsection{Hyperparameters}

\begin{table}[h]
\centering
\small
\begin{tabular}{ll}
\toprule
\textbf{Parameter} & \textbf{Value} \\
\midrule
Base learning rate & $2 \times 10^{-4}$ \\
Batch size & 4 \\
Gradient accumulation steps & 4 \\
Effective batch size & 16 \\
Epochs & 3 \\
Warmup ratio & 0.03 \\
Weight decay & 0.01 \\
Optimizer & AdamW \\
LR scheduler & Cosine \\
Max sequence length & 4096 \\
\midrule
LoRA rank & 16 \\
LoRA alpha & 32 \\
LoRA dropout & 0.1 \\
Target modules & q\_proj, k\_proj, v\_proj, o\_proj \\
\midrule
Quantization & 4-bit NF4 \\
Compute dtype & bfloat16 \\
\bottomrule
\end{tabular}
\caption{Training hyperparameters used for all experiments.}
\label{tab:hyperparameters}
\end{table}

\subsection{Compute Resources}

All experiments were conducted on NVIDIA A100 GPUs (40GB). Training time per configuration: approximately 30 minutes. Total compute: approximately 50 GPU-hours for all experiments.

\section{SafeDialBench Statistics}
\label{sec:appendix-safedialbench}

\begin{table}[h]
\centering
\small
\begin{tabular}{lc}
\toprule
\textbf{Category} & \textbf{Samples} \\
\midrule
Violence & 412 \\
Fraud/Deception & 389 \\
Illegal Activities & 356 \\
Hate/Harassment & 298 \\
Sexual Content & 312 \\
Self-Harm & 270 \\
\midrule
\textbf{Total} & 2,037 \\
\bottomrule
\end{tabular}
\caption{SafeDialBench category distribution.}
\label{tab:safedialbench-stats}
\end{table}

\end{document}